\def\year{2019}\relax
\documentclass[letterpaper]{article} 
\usepackage{aaai19}  
\usepackage{times}  
\usepackage{helvet}  
\usepackage{courier}  
\usepackage{url}  
\usepackage{graphicx}  

\usepackage{booktabs}
\usepackage{multirow}
\usepackage{verbatim}
\title{Predicting the Argumenthood of English Prepositional Phrases}
\author{Najoung Kim,$^\dagger$ Kyle Rawlins,$^\dagger$ Benjamin Van Durme,$^{\dagger,\partial}$ and Paul Smolensky$^{\dagger,\Delta}$\\
$^\dagger$Department of Cognitive Science, Johns Hopkins University, Baltimore, MD, USA\\ 
$^\partial$Department of Computer Science, Johns Hopkins University, Baltimore, MD, USA \\
$^\Delta$Microsoft Research AI, Redmond, WA, USA\\
\url{{n.kim, kgr, vandurme, smolensky}@jhu.edu}}

\begin{document}
\maketitle
\begin{abstract}
Distinguishing between arguments and adjuncts of a verb is a longstanding, nontrivial problem. In natural language processing, argumenthood information is important in tasks such as semantic role labeling (SRL) and prepositional phrase (PP) attachment disambiguation. In theoretical linguistics, many diagnostic tests for argumenthood exist but they often yield conflicting and potentially gradient results. This is especially the case for syntactically oblique items such as PPs. We propose two PP argumenthood prediction tasks branching from these two motivations: (1) binary argument-adjunct classification of PPs in VerbNet, and (2) gradient argumenthood prediction using human judgments as gold standard, and report results from prediction models that use pretrained word embeddings and other linguistically informed features. Our best results on each task are (1) $acc.=0.955$, $F_1=0.954$ (ELMo+BiLSTM) and (2) Pearson's $r=0.624$ (word2vec+MLP). Furthermore, we demonstrate the utility of argumenthood prediction in improving sentence representations via performance gains on SRL when a sentence encoder is pretrained with our tasks.
\end{abstract}

\section{Introduction}

In theoretical linguistics, a formal distinction is made between \textit{arguments} and \textit{adjuncts} of a verb. For example, in the following example, \textit{the window} is an argument of the verb \textit{open}, whereas \textit{this morning} and \textit{with Mary} are adjuncts.

\begin{quote}
	John \textbf{opened} [the window] [this morning] [with Mary].
\end{quote}

\noindent What distinguishes arguments (or \textit{complements}) from adjuncts (or \textit{modifiers})\footnote{Various combinations of the terminology are found in the literature, with subtle domain-specific preferences. We use \textit{arguments} and \textit{adjuncts}, with a rough definition of \textit{arguments} as elements specifically selected or subcategorized by the verb. We use the umbrella term \textit{(verbal) dependents} to refer to both.}, and why is this distinction important? Theoretically, the distinct representations given to arguments and adjuncts manifest in different formal behaviors \cite{chomsky1993lectures,steedman2000syntactic}. There is also a range of psycholinguistic evidence which supports the psychological reality of the distinction \cite{tutunjian2008we}. In natural language processing (NLP), argumenthood information is useful in various applied tasks such as automatic parsing \cite{briscoe1998can} and PP attachment disambiguation \cite{merlo2006notion}. In particular, automatic distinction of argumenthood could prove useful in improving structure-aware semantic role labeling, which has been shown to outperform structure-agnostic models in recent works \cite{marcheggiani2017encoding}. However, argument-adjunct distinction is one of the most difficult linguistic properties to annotate, and has remained unmarked in popular resources including the Penn TreeBank. PropBank \cite{palmer2005proposition} addresses this issue to an extent by providing \textsc{arg-n} labels, but does not provide full coverage \cite{hockenmaier2002acquiring}. Thus, there are theoretical and practical motivations to a systematic approach for predicting argumenthood. 

We focus on PPs in this paper, which are known to be one of the most challenging verbal dependents to classify correctly \cite{abend2010fully}. The paper is structured as follows. First, we discuss the theoretical and practical motivations for PP argumenthood prediction in more detail and review related works. Second, we formulate two different argumenthood tasks---binary and gradient---and describe how each dataset is constructed. Results for each task using various word embeddings and linguistic features as predictors are reported. Finally, we investigate whether better PP argumenthood prediction is indeed useful for NLP. Through a controlled evaluation setup, we demonstrate that pretraining sentence encoders on our proposed tasks improves the quality of learned representations.

\section{Argumenthood Prediction}
\label{theoretical}

\paragraph{Theoretical Motivation} Although arguments and adjuncts are theoretically and practically important concepts, distinguishing arguments from adjuncts in practice is not a trivial problem even for linguists \cite{schutze1995pp}. Numerous diagnostic tests have been proposed in the literature; for instance, omissibility and iterability tests are commonly used \cite{pollard1987information}. However, none of the existing diagnostic tests (or a set of tests) provide necessary or sufficient criteria to determine the status of a verbal dependent. Moreover, it has long been noted that argumenthood is a gradient phenomenon rather than a strict dichotomy (e.g., some arguments are less argument-like than others, resulting in different syntactic and semantic behaviors \cite{rissman2015using}). This raises many interesting theoretical questions such as what kinds of lexical and contextual information affect these judgments, and whether the judgments would be predictable in a principled way given that information. By building prediction models for gradient judgments using lexical features, we hope to gain insights about what factors explain gradience and to what degree they do so.

\paragraph{Utility in NLP} Automatic parsing will likely benefit from argumenthood information. For instance, the issue of PP attachment negatively affects the accuracy of a competitive dependency parser \cite{dasigi2017ontology}. It has been shown that reducing PP attachment errors leads to higher parsing accuracy \cite{agirre2008improving,belinkov2014exploring}, and also that argument-adjunct distinction is useful for PP attachment disambiguation \cite{merlo2006notion}. 

Argument-adjunct distinction is also closely connected to Semantic Role Labeling (SRL). \citeauthor{he2017deep} (\citeyear{he2017deep}) report that even state-of-the-art deep models for SRL still suffer from argument-adjunct distinction errors as well as PP attachment errors. They also observe that errors in widely-used automatic parsers pose challenges to improving performance in syntax-aware neural models \cite{marcheggiani2017encoding}. This suggests that improving parsers with better argumenthood distinction would lead to better SRL performance.

\citeauthor{przepiorkowski2018arguments} (\citeyear{przepiorkowski2018arguments}) discuss the issue of core-noncore distinction being confounded with argument-adjunct distinction in the annotation protocol of Universal Dependencies (UD), which leads to internal inconsistencies. They point out the flaws of the argument-adjunct distinction and suggest a solution that disentangles it better from the core-noncore distinction that UD advocates. This does improve within-UD consistency, but argument-adjunct status is still explicitly encoded in many formal grammars including Combinatory Categorial Grammar (CCG) \cite{steedman2000syntactic}. Thus, being able to predict argumenthood would still be important in improving the quality of resources being ported between different grammars, such as CCGBank. 

\paragraph{Related Work}
Our tasks share a similar objective with \citeauthor{villavicencio2002learning} (\citeyear{villavicencio2002learning}), which is to distinguish PP arguments from adjuncts by an informed selection of linguistic features. However, we do not use logical forms or explicit formal grammar in our models, although the use of distributional word representations may capture some syntactic information. The scale and data collection procedure of our binary classification task (Experiment 1) are more comparable to those of \citeauthor{merlo2006notion} (\citeyear{merlo2006notion}) or \citeauthor{belinkov2014exploring} (\citeyear{belinkov2014exploring}), where the authors construct a PP attachment database from Penn TreeBank data. Our binary classfication dataset is similar in scale, but is based on VerbNet \cite{verbnet} frames. Experiment 2, which is a smaller-scale experiment on predicting gradient argumenthood judgment data from humans, is a novel task to the extent of our knowledge. The crowdsourcing protocol for collecting human judgments is inspired by \citeauthor{rissman2015using} (\citeyear{rissman2015using}).

Our evaluation setup to measure downstream task performance gains from PP argumenthood information (Section~\ref{usefulness}) is inspired by a recent line of efforts on \textit{probing} for evaluating linguistic representations encoded by neural networks \cite{gulordava2018colorless,ettinger2018assessing}. In order to investigate whether PP argument-adjunct distinction tasks have a practical application, we attempt to improve performances on existing tasks such as SRL with an ultimate goal of making sentence representations better and more generalizable (as opposed to representations optimized for one specific task). We use the setup of \textit{pre}training a sentence encoder with a linguistic task of interest, fixing the encoder weights and then training a classifier for tasks other than the pretraining task, using the representations from the frozen encoder \cite{bowman2018looking}. This enables us to compare the utility of information from different pretraining tasks (e.g., PP argument-adjunct distinction) on another task (e.g., SRL) or even multiple external tasks.

\section{Exp. 1: Binary Classification}

\subsection{Task Formulation} 
\label{ex1:data}

\begin{table*}[h]
	\centering
	\begin{tabular}{@{}lllllllllll@{}}
		\toprule
		& \multicolumn{2}{l}{w2v} &\multicolumn{2}{l}{Glove}& \multicolumn{2}{l}{fastText} & \multicolumn{2}{l}{ELMo} \\ \midrule
		Classification model       & Acc. & $F_1$ & Acc. & $F_1$ & Acc. & $F_1$ & Acc. & $F_1$   \\ \midrule
		BiLSTM + MLP &   94.0  &  94.0 \hspace{0.8cm} &  94.5 & 94.4 \hspace{0.8cm} & 94.6 & 94.6 \hspace{0.8cm} & \textbf{95.5}& \textbf{95.4} \\
		Concatenation + MLP & 92.4 & 92.5 & 93.3& 93.3& 93.6 & 93.5 & 94.4 & 94.4\\
		BoW + MLP & 91.9& 91.9 & 92.4 &92.4 & 92.7& 92.5& 93.7 & 93.7\\
		Majority class (== chance)       & 50.0      & 50.0 &  50.0 & 50.0  & 50.0 &50.0 & 50.0 & 50.0\\
		\bottomrule
	\end{tabular}
	\caption{Test set performance on the binary classification task  ($n=4064$).}
	\label{ex1:accuracy}
\end{table*}

\paragraph{Class Labels} We use VerbNet subcategorization frames to define the argument-adjunct status of a verb-PP combination. This means if a certain PP is listed as a subcategorization frame under a certain verb entry, the PP is considered to be an argument of the verb. If it is not listed as a subcategorization frame, it is considered an adjunct of the verb. This way of defining argumenthood has been proposed and studied by \citeauthor{mcconville2008evaluating} (\citeyear{mcconville2008evaluating}). Their evaluation suggests that even though VerbNet is not an exhaustive list of subcategorization frames and not all frames listed are strictly arguments, VerbNet membership is a reasonable proxy of PP argumenthood. We chose VerbNet over PropBank \textsc{arg-n} and \textsc{am} labels, which are also a viable proxy for argumenthood \cite{abend2010fully}, since VerbNet's design goal of exhaustively listing frames for each verb better matches our task that requires broad type-level coverage of V-PP constructions. 

\begin{figure}[h]
\centering
	\includegraphics[width=0.9\linewidth]{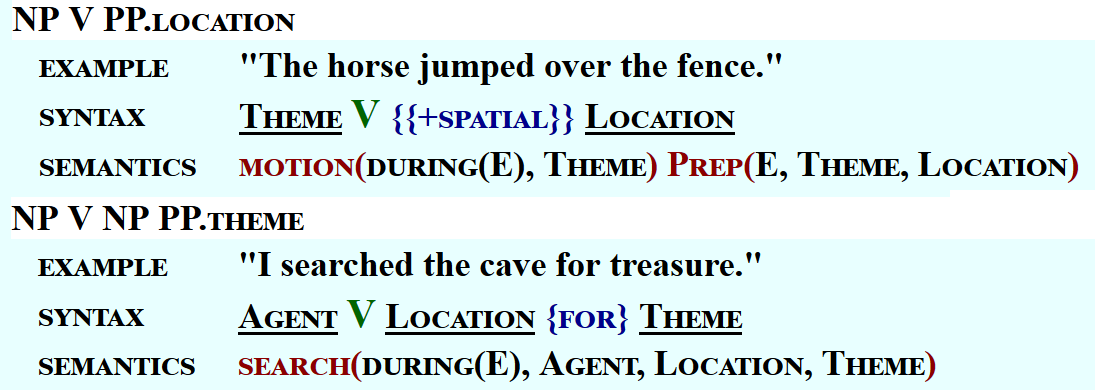}
	\caption{Examples of \textsc{PP.x} frames in VerbNet.}
	\label{fig:ppframes}
\end{figure}

\paragraph{Verbs and Prepositions} 2714 unique verbs (V) and 60 unique prepositions (P) are used to generate all possible combinations of \{\textsc{V, P}\}. These are all unique verb entries that pertain to a single set of VerbNet class and all prepositions that appear in VerbNet PP frames excluding multi-word prepositions (e.g., \textit{all over, on top of}). Some frames are only defined by features such as \textsc{\{+spatial\}}, without specifying which specific prepositions these features correspond to (see Figure~\ref{fig:ppframes}). For such featurally-marked frames, a manual mapping was made to preposition sets constructed approximately based on PrepWiki sense annotations \cite{schneider2015hierarchy}. As a result, we have $2714*60\textnormal{ = }162,480$ different \textsc{\{V, P\}} tuples that either correspond to ($\mapsto$\textsc{arg}) or do not correspond to ($\mapsto$\textsc{adj}) a subcategorization frame.

\paragraph{Dataset and Task Objective}  Since each verb subcategorizes only a handful of prepositions out of the possible 60, the distribution of labels \textsc{arg} and \textsc{adj} is heavily skewed towards \textsc{adj}. The ratio of \textsc{arg}:\textsc{adj} labels in the whole dataset is approximately 1:10. For this reason, we use randomized subsampling of the negative cases to construct a balanced dataset. Since there were $13,544$ datapoints with label 1 in the whole set, the same number of label 0 datapoints were randomly subsampled. This balanced dataset ($n=27,088$) is randomly split into 70:15:15 train:dev:test sets.

The task is to predict whether a given \textsc{\{V, P\}} pair is an argument or an adjunct construction (i.e., whether it is an existing VerbNet subcategorization frame or not). Performance is measured by classification accuracy and $F_1$ on the test set. 

\paragraph{Full-sentence Tasks} The meaning of the complement noun phrase (NP) of the preposition can also be a clue to determining argumenthood. We did not include the NP as a part of the input in our main task dataset because the argumenthood labels we are using are type-level and not labels for individual instantiation of the types (token-level). However,the NP meanings, especially thematic roles, do play crucial roles in argumenthood. To address this concern, we propose variants of the main task that give prediction models access to information from the NP by providing full sentence inputs. We report additional results on one of the full-sentence variants (ternary classification) using two of the best-performing model setups for the main task (Table~\ref{ex1-2:full sentence task}).

The full-sentence variants of the main task dataset are constructed by performing a heuristic search through the Stanford Natural Language Inference (SNLI; \cite{bowman2015large}) and Multi-genre Natural Language Inference (MNLI; \cite{williams2018broad}) datasets using the syntactic parse trees provided, to find sentences that contain a particular \{\textsc{V, P}\} construction. Note that the full sentence data is noisier compared to the main dataset (1) because the trees are parser outputs and (2) the original type-level gold labels given to the \{\textsc{V, P}\} were unchanged regardless of what the NP may be. Duplicate entries for the same \{\textsc{V, P}\} were permitted as long as the sentences themselves were different. In the first task variant, for case where no examples of the input pair is found in the dataset, we retained the original label. In the second variant, we assign a new \textsc{unobserved} label to such cases. This helps filter out overgenerated adjunct labels in the original dataset, where the \{\textsc{V, P}\} is not listed as a frame because it is an impossible or an infelicitous construction. It also eliminates the need for subsampling, since the three labels were reasonably balanced. 

We chose NLI datasets as the source of full sentence inputs over other parsed corpora such as the Penn TreeBank for the following two reasons. First, we wanted to avoid using the the same source text as several downstream tasks we test in Section~\ref{usefulness} (e.g., CoNLL-2005 SRL \cite{carreras2005introduction} uses sections of the Penn TreeBank), in order to separate out the benefits of seeing the source text at train time from the benefits of structural knowledge gained from learning to distinguish PP arguments and adjuncts. Second, we wanted both simple, short sentences (SNLI) and complex, naturally-occuring sentences (MNLI) from datasets of a consistent structure.

\subsection{Model}
\label{ex1:model}
\paragraph{Input Representation} We report results using 4 different types of word embeddings (word2vec \cite{mikolov2013efficient}, GloVe \cite{pennington2014glove}, fastText \cite{bojanowski2016enriching}, ELMo \cite{peters2018deep}) to represent the input tuples. Publicly available pretrained embeddings provided by the respective authors\footnote{\url{code.google.com/archive/p/word2vec/}\\ \url{nlp.stanford.edu/projects/glove/}\\
\url{github.com/facebookresearch/fastText} \\
\url{github.com/allenai/allennlp}} are used.

\begin{table*}[h]
	\centering
	\begin{tabular}{@{}lllllllllll@{}}
		\toprule
		& \multicolumn{2}{l}{w2v}  & \multicolumn{2}{l}{Glove} &  \multicolumn{2}{l}{fastText} & \multicolumn{2}{l}{ELMo}  \\ \midrule
		Classification model       & Acc. & $F_1$ & Acc. & $F_1$ & Acc. & $F_1$ & Acc. & $F_1$   \\ \midrule
		BiLSTM + MLP &  95.6  & 94.3 \hspace{1cm} & 97.0  & 96.1  \hspace{1cm} & 96.9 & 96.0  \hspace{1cm} & \textbf{97.4} & \textbf{96.6} \\
		BoW + MLP & 93.8 & 92.0& 93.7& 91.8& 94.2 & 92.5& 95.9 & 94.7 \\
		Majority class & 38.8 & 38.8 & 38.8 & 38.8 & 38.8 & 38.8 & 38.8 & 38.8 \\ 
		Chance       & 33.3  & 33.3  & 33.3 & 33.3 & 33.3 & 33.3 & 33.3  & 33.3  \\
		\bottomrule
	\end{tabular}
	\caption{Model performances on \textbf{full-sentence, unobserved label included} variant of the original classification task (now ternary classification) on the test set ($n=18,764$).}
	\label{ex1-2:full sentence task}
\end{table*}

\paragraph{Classifiers}
Our current best model uses a combination of bidirectional LSTM (BiLSTM) and multi-layer perceptron (MLP) for the binary classification task. We first obtain a representation of the given \{\textsc{V, P}\} sequence using an encoder and then train an MLP classifier (Eq. 1) on top to perform the actual classification. The BiLSTM encoder is implemented using the AllenNLP toolkit \cite{gardner2018allennlp}, and the MLP classifier is a simple feedforward neural network with a single hidden layer that consists of $512$ units (Eq. 1). We also test models that use the same MLP classifier with concatenated input vectors (Concatenation + MLP) or bag-of-words encodings of the input (BoW + MLP).

\begin{equation}
\label{eq1}
\fontsize{7.5pt}{8.5pt}\selectfont
\begin{array}{ll}
l = argmax(\sigma(W_o [ \tanh(W_h[V_{vp}] + b_h)] + b_o))
\end{array}
\end{equation}

\noindent $V_{vp}$ is a linear projection ($d=512$) of the output of an encoder (encoder states followed by max-pooling or a concatenated word vector), and $\sigma$ is the softmax activation function. The output of Eq.~\ref{eq1} is the label (1 or 0) that is more likely given $V_{vp}$. The models are trained using Adadelta \cite{zeiler2012adadelta} with cross-entropy loss and batch size $=32$. Several other off-the-shelf implementations from the Python library \textit{scikit-learn} were also tested in place of MLP, but since MLP consistently outperformed other classifiers, we only list models that use MLP classifiers out of the models we tested (Table~\ref{ex1:accuracy}).

\subsection{Results}
Table~\ref{ex1:accuracy} compares the performance of the tested models. The most trivial baseline is chance-level classification, which would yield 50\% accuracy and $F_1$. Since the labels in the dataset are perfectly balanced by randomized subsampling, majority class classification is equivalent to chance-level classification. All nontrivial models outperform chance, and out of all models tested ELMo yielded the best performance. Using concatenated inputs that preserve the linear order of the inputs increases both accuracy and $F_1$ by around 1.5\%p over bag-of-words, and using a BiLSTM to encode the V+P representation adds 1.5\%p improvement across the board.

We additionally report results on the full-sentence, ternary-classification variant of the task (description in Section~\ref{ex1:data}) from the best model on the main task. The results are given in Table~\ref{ex1-2:full sentence task}. We did not test the concatenation model since the dimensionality of the vectors would be too large with full sentence inputs. All tested models perform over chance, with ELMo achieving the best performance once again. We observe a similar gain of around 3\%p by replacing BoW with BiLSTM as in the main task.

\section{Exp. 2: Gradient Argumenthood Prediction}
As discussed in Section~\ref{theoretical}, there is much work in theoretical linguistics literature that suggests argument-adjuncthood is a continuum rather than a dichotomy. We propose a gradient argumenthood prediction task and test regression models that use a combination of embeddings and lexical features. Since there is no publicly available gradient argumenthood data, we collected our own dataset via crowdsourcing\footnote{See Supplemental Material for examples of questions given to participants. Detailed protocol and theoretical analysis of the data are omitted; it will be discussed in a separate theoretically-oriented paper in preparation.}. Due to the resource-consuming nature of human judgment collection, the size of the dataset is limited ($n=305$, 25-way redundant). This task serves as a small-scale, proof-of-concept test for whether a reasonable prediction of gradient argumenthood judgments is possible with informed selection of lexical features (and if so, how well different models perform and what features are informative).

\begin{table*}[h]
	\centering
	\begin{tabular}{@{}llllllllll@{}}
		\toprule
		 all $d=300$ & \multicolumn{3}{l}{w2v-googlenews} & \multicolumn{3}{l}{GloVe} & \multicolumn{3}{l}{fastText}\\ \midrule
		Model       & Pearson's $r$ & $R^2$ & $R^2_{adj}$ & Pearson's $r$ & $R^2$ & $R^2_{adj}$ & Pearson's $r$ & $R^2$ & $R^2_{adj}$  \\ \midrule
		Simple MLP & 0.554 & 0.255 &  0.231  \hspace{1cm} & 0.568 & 0.268 & 0.245 \hspace{1cm} & 0.582 & 0.281 & 0.257\\
		Linear & 0.579 & 0.280 & 0.257 & 0.560 & 0.243 & 0.219 & 0.591 & 0.291 & 0.268 \\				
		SVM & 0.561 & 0.243 & 0.218 & 0.463 & 0.110 & 0.082 & 0.580 & 0.267 & 0.243 \\
		\midrule
	\end{tabular}	
	\begin{tabular}{@{}llllllllll@{}}
		\midrule
		$d \ge 1000$& \multicolumn{3}{l}{w2v-wiki\footnotemark \hspace{0.05cm} ($d=1000$)}  & \multicolumn{3}{l}{ELMo ($d=1024$)} \\ \midrule
		Model       & Pearson's $r$ & $R^2$ & $R^2_{adj}$ & Pearson's $r$ & $R^2$  & $R^2_{adj}$ \\ \midrule
		Simple MLP & \textbf{0.624}  & \textbf{0.330} &  \textbf{0.309} \hspace{1cm} & 0.609 & 0.304 & 0.281 \\
		Linear & 0.609 & 0.311 & 0.289 & 0.586 & 0.293  & 0.270 \\	
		SVM & 0.549 & 0.237 & 0.213 & 0.337 & 0.052  & 0.022  \\
		\bottomrule
	\end{tabular}	
	\caption{10-fold cross-validation results on the gradient argumenthood prediction task.}
	\label{ex2:results}
\end{table*}
\footnotetext{\url{github.com/idio/wiki2vec}}

\begin{figure}[h]
	\centering
	\includegraphics[width=0.8\linewidth]{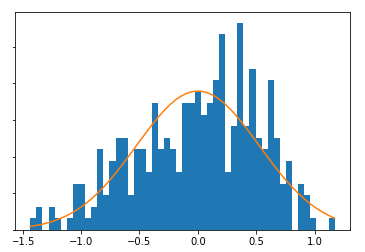}
	\caption{Distribution of argumenthood scores in our gradient argumenthood dataset.}
	\label{fig:dist}
\end{figure}

\subsection{Data}
The gradient argumenthood judgment dataset consists of 305 sentences that contain a single main verb and a PP dependent of that verb. All sentences are adaptations from example sentences in VerbNet PP subcategorization frames or sentences containing \textsc{arg-n} PPs in PropBank. To capture a full range of the argumenthood spectrum from fully argument-like to fully adjunct-like, we manually augmented the dataset by adding more strongly adjunct-like examples. These examples are generated by substituting the PP of the original sentence with a felicitous adjunct. This step is necessary since PPs listed as a subcategorization frame or \textsc{arg-n} are more argument-like, as discussed in Section~\ref{ex1:data}. 

25 participants were recruited to produce argumenthood judgments about these sentences on a 7-point Likert scale, using protocols adapted from a prior work for collecting similar judgments \cite{rissman2015using}. Larger numbers are given an interpretation of being more argument-like and smaller numbers, more adjunct-like. The results were \textit{z-}normalized within-subject to adjust for individual differences in the use of the scale, and then the final argumenthood scores were computed by averaging the normalized scores across all participants. The result is a set of values on a continuum, each value representing each sentence in the dataset. The actual numbers range between $[-1.435, 1.172]$, roughly centered around zero ($\mu=1.967e\textnormal{-}11, \sigma=0.526$). The distribution is slightly left-skewed (right-leaning), with more positive values than negative values ($z=2.877, p<.01$; see Figure~\ref{fig:dist}). Here are some examples with varying degrees of argumenthood, indicated by the numbers:

\begin{quote}
	\textit{I \textbf{whipped} the sugar [with cream].} (0.35)\\
	\hspace{-0.5cm}\textit{The witch \textbf{turned} him [into a frog].} ($0.57$) \\
	\hspace{-0.5cm}\textit{The children \textbf{hid} [in a hurry].} ($-0.41$) \\
	\hspace{-0.5cm}\textit{It \textbf{clamped} [on his ankle].} ($0.66$) \\
	\hspace{-0.5cm}\textit{Amanda \textbf{shuttled} the children [from home].} ($-0.1$)
\end{quote}

\noindent We refrain from assigning definitive interpretations to the absolute values of the scores, but how the scores compare to each other gives us insight into the relative difference in argumenthood. For instance, \textit{shuttled from home} with a score of $-0.1$ is more adjunct-like than a higher-scoring construction such as \textit{clamped on his ankle} $(0.66)$, but more argument-like compared to \textit{hid in a hurry} with a score of $-0.41$. This matches the intuition that a locative PP \textit{from home} would be more argument-like to a change-of-location predicate \textit{shuttle}, compared to a manner PP like \textit{in a hurry}. However, it is still less argument-like than a more clearly argument-like PP \textit{on his ankle} that is selected by \textit{clamp}.

\subsection{Model}
\label{ex2:model}

\paragraph{Features}
We use the same sets of word embeddings we used in Experiment 1 as base features, but we reduced their dimensionality to $5$ via Principal Component Analysis. This reduction step is necessary due to the large dimensionality ($d \ge 300$) of the word vectors compared to the small size of our dataset ($n=305$).  Various features in addition to the embeddings of verbs and prepositions were also tested. The features we experimented with include semantic proto-role property scores \cite{reisinger2015semantic} of the target PP (normalized mean across 5 annotators), mutual information (MI) \cite{aldezabal2002learning}, word embeddings of the nominal head token of the NP under the PP in question, existence of a direct object, and various interaction terms between the features (e.g., additive, subtractive, inner/outer products). The following lexical features were selected for the final models based on dev set performance: embeddings of the verb, preposition, nominal head, mutual information and existence of a direct object (D.O.). The intuition behind including D.O. is that if there exists a direct object in the given sentence, the syntactically oblique PP dependent would seem comparatively less argument-like compared to the direct object. This feature is expected to reduce the noise introduced by different argument structures of the main verbs.

\begin{table*}[h]
	\centering
	\begin{tabular}{@{}lllllllllll@{}}
		\toprule
		& \multicolumn{3}{l}{w2v-wiki} & \multicolumn{3}{l}{ELMo} & \multicolumn{3}{l}{No embeddings}\\ \midrule
		Model       & Pearson's $r$ & $R^2$ & $R^2_{adj}$ & Pearson's $r$ & $R^2$ & $R^2_{adj}$ & Pearson's $r$  & $R^2$ & $R^2_{adj}$  \\ \midrule
		Embeddings &  0.430 & 0.064 & 0.046 \hspace{0.7cm} & 0.404 &  0.063 & 0.044 \hspace{0.7cm} & - & - & -\\
		{} {} {} {} +MI & 0.464 & 0.158  & 0.138 & 0.458 &  0.153 & 0.133 & 0.376 & 0.083 & 0.079 \\
		{} {} {} {} +D.O.  & 0.575 & 0.245 & 0.227 & 0.530 & 0.202 & 0.183 & 0.301 & 0.029 & 0.025\\
		{} {} {} {} +diag. & 0.449 & 0.125 & 0.104 & 0.440 & 0.138 & 0.117 & 0.268 & 0.027 & 0.023 \\
		{} {} {} {} +MI. +D.O.  & 0.586 & 0.278 & 0.258 & 0.607 & 0.297 & 0.277 & 0.466 & 0.165 & 0.158 \\
		{} {} {} {} +MI. +diag. & 0.515 & 0.205 & 0.182 & 0.512 & 0.193 & 0.170 & 0.436 & 0.141 & 0.134  \\
		{} {} {} {} +diag. +D.O. & 0.572 & 0.265 & 0.245 & 0.535 & 0.224 & 0.203 & 0.392 & 0.114 & 0.107 \\
		{} {} {} {} +all & 0.624 & 0.330  & 0.309 & 0.609 & 0.304 & 0.281
		&  0.516 & 0.215 & 0.206 \\
		\bottomrule
	\end{tabular}
	\caption{Ablation results from the two best models and a non-embedding features only-model (10-fold cross validation).}
	\label{ex2:ablation-2}
\end{table*}

We also include a \textit{diagnostics} feature which is a weighted combination of two different traditional diagnostic test results (omissibility and pseudo-cleftability) produced by a linguist with expertise in theoretical syntax. Unlike all other features in our feature set, this diagnostics feature is not straightforwardly computable from corpus data. We add this feature in order to examine how powerful traditional linguistic diagnostics are in capturing gradient argumenthood.

\paragraph{Regression Model}
The selected features are given as inputs to an MLP which is equivalent to Eq. 1 in Experiment 1 except that it outputs a continuous value. This regressor consists of an input layer with $n$ units (corresponding to $n$ features), \textit{m} hidden units and a single output unit. Smooth L1 loss is used in order to reduce sensitivity to outliers, and \textit{m}, the activation function (ReLU, Tanh or Sigmoid), optimizers and learning rates are all tuned using the development set for each individual model. We limit ourselves to a simpler MLP-only model for this experiment; the BiLSTM encoder model suffered from overfitting.

\paragraph{Evaluation Metrics}
We use 15\% of the dataset as development set, and train/test using 10-fold cross-validation on the remaining 85\% rather than reporting performance on a fixed test split. This is because the credibility of performance on one test split may be questioned due to the small sample size. Pearson's $r$ averaged over the 10 folds using Fisher \textit{z}-transformation is the main metric. Mean $R^2$ and Adjusted $R^2$ ($R^2_{adj}$) are also reported to account for the potentially differing number of predictors in the ablation experiment.

\begin{table*}[h]
    \centering
	\begin{tabular}{@{}l|l|llll@{}}
		\toprule
		& & \multicolumn{4}{c}{Sentence encoder pretraining tasks} \\ 
		Test tasks & metric & Random & Arg & Arg fullsent & Arg fullsent 3-way \\ \midrule
		SRL-CoNLL2005 (WSJ)  & $F_1$ &  81.7 & \textbf{83.9$^{***}$} & \textbf{84.7$^{***}$} & \textbf{84.5$^{***}$}  \\
		SRL-CoNLL2012 (OntoNotes) & $F_1$  & 77.3 & \textbf{80.2$^{***}$} & \textbf{80.4$^{***}$} & \textbf{80.7$^{***}$} \\
	    PP attachment \cite{belinkov2014exploring} & $acc.$ & 87.5 & 87.6 & 88.2 & 87.0 \\	     
		\bottomrule
	\end{tabular}
\caption{\normalsize Gains over random initialization from pretraining sentence encoders on PP argumenthood tasks. ($^{***}: p<.001$)}
\label{exp:pretrain}
\end{table*}

\subsection{Results and Discussion}

Table~\ref{ex2:results} reports performances on the regression task. Results from several off-the-shelf regressors are reported for comparison. ELMo embeddings again produced the best results among the embeddings used in Experiment 1 (although our ablation study reveals that using only word embeddings as predictors is not sufficient to obtain satisfactory results). We further speculated that the dimensionality of the embeddings may have impacted the results, and ran additional experiments using higher-dimensional embeddings that were publicly available. Higher-dimensional embeddings did indeed lead to performance improvements, even though the actual inputs given to the models were all PCA-reduced to $d=5$. From this observation, we could further improve upon the inital ELMo results. Results from the best model (w2v-wiki) are given in addition to the set of results using the same embeddings as the models in Experiment 1. This model uses 1000-$d$ word2vec features with additional interaction features (multiplicative, subtractive) that improved dev set performance. 

\paragraph{Ablation}
We conducted ablation experiments with the two best-performing models to examine the contribution of non-embedding features discussed in Section~\ref{ex2:model}. Table~\ref{ex2:ablation-2} indicates that any linguistic feature contributes positively towards performance, with the direct object feature helping both word2vec and ELMo models the most. This supports our initial hypothesis that  adding the direct object feature would help reduce noise in the data. When only the linguistic features are used without embeddings as base features, mutual information is the most informative. This suggests that there is some (but not complete) redundancy in information captured by word embeddings and mutual information. The diagnostics feature is informative but is a comparatively weak predictor, which aligns with the current state of diagnostic acceptability tests---they are sometimes useful but not always, especially with respect to syntactically oblique items such as PPs. This behavior of the diagnostics predictor adds credibility to our data collection protocol.

\section{Why Is This a Useful Standalone Task?}
\label{usefulness}
In motivating our tasks, we suggested that PP argumenthood information could improve existing NLP task performance such as SRL and parsing. We investigate whether this is a grounded claim by testing two separate hypotheses: (1) whether the task is indeed useful, and if so, (2) whether it is useful as a standalone task. We leave the issue of gradient argumenthood to future work for now, since the dataset is currently small and the notion of gradient argumenthood is not yet compatible with formulations of many NLP tasks.

\subsection{Improving Representations with Pretraining}
We first test the utility of the binary argumenthood task in improving performances on existing NLP tasks. We selected three tasks that may benefit from PP argumenthood information: SRL on Wall Street Journal (WSJ) data (CoNLL 2005; \citeauthor{carreras2005introduction} \citeyear{carreras2005introduction}), SRL on OntoNotes Corpus (CoNLL 2012 data; \citeauthor{pradhan2012conll} \citeyear{pradhan2012conll})\footnote{Tasks are labeling only, as described in \citeauthor{tenney2019what} (\citeyear{tenney2019what}).}, and PP attachment disambiguation on WSJ \cite{belinkov2014exploring}.

We follow \citeauthor{bowman2018looking} (\citeyear{bowman2018looking})'s setup to pretrain and evaluate sentence encoders\footnote{\url{github.com/jsalt18-sentence-repl/jiant}}. If learning to make correct PP argumenthood distinction teaches models knowledge that is generalizable to the new tasks, the classifier trained on top of the fixed-weights encoder will perform better on those tasks compared to a classifier trained on top of an encoder with randomly initialized weights. Improvements over the randomly initialized setup from pretraining on our main PP argumenthood task (Arg) and its full-sentence variants (Arg fullsent and Arg fullsent 3-way; see Section~\ref{ex1:data} for details) are shown in Table~\ref{exp:pretrain}. Only statistically significant ($p<.05$) improvements over the random encoder model are bolded, with significance levels calculated via Approximate Randomization \cite{yeh2000more} ($R=1000$). The models trained on PP argumenthood tasks perform significantly better than the random initalization model in both SRL tasks, which supports our initial claim that argumenthood tasks can be useful for SRL. Although not all errors made by the models were interpretable, we found interesting improvements such as the model trained on the PP argumenthood task being slightly more accurate than the random initialization model on \textsc{AM-DIR, AM-LOC}, and \textsc{AM-MNR} labels.

However, we did not observe significant improvements for the PP attachment disambiguation task. We speculate that since the task as formulated in \citeauthor{belinkov2014exploring} (\citeyear{belinkov2014exploring}) requires the model to understand PP dependents of NPs as well as VPs, our tasks that focus on verbal dependents may not provide the full set of linguistic knowledge necessary to solve this task. Nevertheless, our models are not significantly worse than the baseline, and the accuracy of the Arg fullsent model (88.2\%) was comparable to a model that uses an encoder directly trained on PP attachment (88.7\%). 

Secondly, we discuss whether it is indeed useful to formulate PP argumenthood prediction as a separate task. The questions that need to be answered are (1) whether it would be the same or better to use a different pretraining task that would provide similar information (e.g., PP attachment disambiguation), and (2) whether the performance gain can be attributed to simply seeing more datapoints at train time rather than to the regularities we hope the models would learn through our task. Table~\ref{exp:comparison} addresses both questions; we compare models pretrained on argumenthood tasks to a model pretrained directly on the PP attachment task listed in Table~\ref{exp:pretrain}. All models trained on PP argumenthood prediction outperform the model trained on PP attachment, despite the fact that the latter has advantage for SRL2005 since the tasks share the same source text (WSJ). Furthermore, the variance in the sizes of the datasets indicates that the reported performance gains cannot solely be due to the increased number of datapoints seen during training.

\begin{table}[h]
	\begin{tabular}{@{}lllll@{}}
		\toprule
		 \hspace{1.2cm} & PP att. & Arg & Arg full & Arg full 3-way  \\ \midrule
		 Size & 32k & 19k & 58k & 87k  \\ \midrule
    	SRL2005   & 80.2  & \textbf{83.9$^{***}$} & \textbf{84.7$^{***}$} & \textbf{84.5$^{***}$}    \\	
		SRL2012   & 79.8    & \textbf{80.2$^{***}$} & \textbf{80.3$^{***}$} & \textbf{80.7$^{***}$}    \\	
		\bottomrule
	\end{tabular}
\caption{\normalsize Comparison against using PP attachment directly as a pretraining task ($^{***}: p<.001$).}
\label{exp:comparison}
\end{table}

\section{Conclusion}
We have proposed two different tasks---binary and gradient---for predicting PP argumenthood, and reported results on each using four different types of word embeddings as base predictors. We obtain 95.5 accuracy and 95.4 $F_1$ in the binary classification task with BiLSTM and ELMo, and $r=0.624$ for the gradient human judgment prediction task. Our overall contribution is threefold: first, we have demonstrated that a principled prediction of both binary and gradient argumenthood judgments is possible with informed selection of lexical features; second, we justified the utility of our binary PP argumenthood classification as a standalone task by reporting performance gains on multiple end-tasks through encoder pretraining. Finally, we have conducted a proof-of-concept study with a novel gradient argumenthood prediction task, paired with a new public dataset\footnote{To be released at: \url{decomp.io}}.

\subsection{Future Work}
The pretraining approach holds promise in understanding and improving neural network models of language. Especially for end-to-end models, this method has an advantage over architecture engineering or hyperparameter tuning in terms of interpretability. That is, we can attribute the source of the performance gain on end tasks to the knowledge necessary to do well on the pretraining task. For instance, in Section~\ref{usefulness} we can infer that that knowing how to make correct PP argumenthood distinction helps models encode representations that are more useful for SRL. Furthermore, we believe it is important to contribute to the recent efforts for designing better probing tasks to understand what machines really know about natural language (as opposed to directly taking downstream performances as metrics of better models). We hope to scale up our preliminary experiments and will continue to work on developing a set of linguistically informed probing and pretraining tasks for higher-quality, better-generalizable sentence representations.

\section*{Acknowledgments}
This research was supported by NSF INSPIRE (BCS-1344269) and DARPA LORELEI. We  thank C. Jane Lutken, Rachel Rudinger, Lilia Rissman, G\'{e}raldine Legendre and the anonymous reviewers for their constructive feedback. Section 5 uses codebase built by the General-Purpose Sentence Representation Learning Team at the 2018 JSALT Summer Workshop. 

\bibliographystyle{aaai}

\fontsize{9.0pt}{10.0pt}\selectfont
\bibliography{aaai.bib}

\end{document}